# A New Approach to Content-based File Type Detection [†]

Mehdi Chehel Amirani       Mohsen Toorani [+]       Ali A. Beheshti

## Abstract

*File type identification and file type clustering may be difficult tasks that have an increasingly importance in the field of computer and network security. Classical methods of file type detection including considering file extensions and magic bytes can be easily spoofed. Content-based file type detection is a newer way that is taken into account recently. In this paper, a new content-based method for the purpose of file type detection and file type clustering is proposed that is based on the PCA and neural networks. The proposed method has a good accuracy and is fast enough.*

## 1. Introduction

With the everyday increasing importance of privacy, security, and wise use of computational resources, the corresponding technologies are increasingly being faced with the problem of file type detection. True identification of computer file types is a difficult task especially when dealing with suspicious goals. Computers are dealing with the huge number of file formats that are transmitting between the insider and outsider networks. Without the true file type detection, the security will not be achievable. File type detection has the most usage and importance in the proper functionality of operating systems, firewalls, intrusion detection systems, anti-viruses, filters, steganalysis, computer forensics, and applications dealing with the file type classification [1]. Although there are many applications dealing with file type detection, they have very limited methods for determining the true type of files. The newest method of file type detection is based on the file contents. McDaniel and Heydari published the first paper on content-based file type detection [2, 3].

They introduced three introductory algorithms for the content-based file type detection: *Byte frequency Analysis* (BFA), *Byte Frequency Cross-correlation* (BFC), and *File Header/Trailer* (FHT) analysis. They deployed such algorithms on the whole content of sample files to produce a *fingerprint* of each file type. Li et al. [4] performed some improvements on the McDaniel and Heydari's method [3] to improve its accuracy. Their improvements was using multi-centroids for some file types, truncating the sample files from their beginning instead of using all the file contents, using K-Means algorithm under Manhattan distance to produce the *fileprints* and deploying Mahalanobis distance for comparison between the unknown samples and fileprints. Dunham et al. [5] deployed neural networks to classify file types of stream ciphers in depth, i.e. the files encrypted with the same key. They used byte frequency, byte frequency of autocorrelation, and 32 bytes of header as the selected features of their test samples. Karresand and Shahmehri [6] used the mean and standard deviation of *Byte Frequency Distribution* (BFD) to model the file types. Their method is based on data fragments of files and does not need any metadata. Zhang et al. [7] used the BFD in conjunction with a simple Manhattan distance comparison to detect whether the examined file is executable or not.

In this paper, a new content-based file type detection method is introduced that deploys the *Principal Component Analysis* (PCA) and unsupervised neural networks for the automatic feature extraction. This paper is organized as follows. Section 2 introduces the file type detection methods. Section 3 briefly describes PCA and the concept of unsupervised neural networks. Our proposed method is presented at Section 4. The simulation results are given in section 5, and section 6 provides the conclusions.

---

## 2. File Type Detection Methods

True identification of a file format is a tedious task. There are catalogues containing several thousand of known file types [8], without having any global standard for the file types. File type detection methods can be categorized into three kinds: extension-based, magic bytes-based, and content-based methods, each of them has its own strengths and weaknesses, and none of them are comprehensive or foolproof enough to satisfy all the requirements.

The fastest and easiest method of file type detection is the extension-based method. The Microsoft's operating systems use such approach almost exclusively. All the file types, at least in the Windows-based systems, are generally accompanied by an extension. This approach can be applied to both binary and text files. While it does not need to open the files, it is by far the fastest way to classify the files. However, it has a great vulnerability while it can be easily spoofed by a simple file renaming.

The second method of file type detection that is devoted to the binary files is based on the magic bytes. The magic bytes are some predefined signatures in the header or trailer of binary files. A file header is the first portion of a computer file that contains metadata. Metadata is the background information that describes the content, quality, condition, and other appropriate characteristics of the data. The file header contains necessary information for the corresponding application to recognize and understand the file. Magic bytes may include some extra information regarding the tool and the tool's version that is used to produce the file. Checking the magic bytes of a file is a little slower than just checking its extension since the file should be opened and its magic bytes should be read. The magic bytes method is taken by many UNIX-based operating systems. However, it has some drawbacks: the magic bytes are not used in all file types. They only work on the binary files and are not an enforced or regulated aspect of the file types. They vary in length for different file types and do not always give a very specific answer. There are several thousands file types for which magic bytes are defined and there are multiple lists of magic bytes that are not completely consistent. Since there are not any standard for what a file can contain, the creators of a new file type usually include something to uniquely identify their file types. Some programs or program developers may never put any magic bytes at the beginning of their file types. This approach can be also spoofed. Altering the magic bytes of a file may not disturb its functionality but can defeat the true file type detection.

The third method of file type detection is to consider the file contents and using the statistical modeling techniques. It is a new research area and is the only way to determine the spoofed file types. It can reveal the malicious file types that their contents do not match with their claimed types. It is based on the byte values inside of different computer files. Each computer byte consists of eight bits so it can accept 256 different values varying between 0 and 255. The BFD of a file can be simply found by reading the contents of that file and counting the number of times that each byte value occurs. It is believed that different files of the same type have the same characteristics, which can be extracted to be used in the file type detection. In this approach, several sample files of the same type is given and a fileprint, something similar to a fingerprint, is produced from the sample files. Whenever an unknown file is examined, its fileprint will be produced with the same process and it will be compared with the collection of previously produced fileprints. To produce a fileprint some features of the file type should be selected and extracted. There are some methods that can be used for the feature extraction. The original principle is to use the BFD of file contents and manipulate with its statistical features. Such statistical measurements together form a model of the chosen file type, sometimes called a centroid. It is also possible to produce several centroids from a file type (multi-centroids). The centroids are then compared to an unknown sample file or data fragment, and the distance between the sample and the centroids is then calculated. If such distance is lower than a predefined threshold, the examined file is categorized as being of the same file type that the centroid represents.

## 3. Feature Extraction

The problem of optimally selecting the statistical features is known as *feature selection* and *feature extraction*. While the feature extraction creates a smaller set of features from linear or nonlinear combinations of the original features, the feature selection chooses a subset of the original features. Feature selection transforms a data space into a feature space that theoretically has the same dimension as the original data space. Such transformation is practically designed in a way that a reduced number of effective features represent the data set while retaining most of intrinsic information content of the data. A reduction in the dimension of the input space is generally accompanied by losing some of information. The goal of dimensionality reduction is to preserve the relevant information as much as possible. In this section, the concepts of PCA and unsupervised neural networks that we use them for the feature extraction is briefly described.

## 3.1. Principle Component Analysis

The PCA is a well-known feature extraction technique in the multivariate analysis. It is an orthogonal transformation of the coordinate in which the data is described. A small number of principle components are usually sufficient to account for the most of structure in the data. Let $\mathbf{X} = \{\mathbf{x}_n \in R^d \mid n = 1,...,N\}$ represents an $d$-dimensional dataset. The PCA tries to find a lower dimensional subspace to describe the original dataset while preserving the information as much as possible, so a new $k$-dimensional dataset $\mathbf{Z} = \{\mathbf{z}_n \in R^k \mid n = 1,...,N\}$ will be produced, where $k$ is smaller than $d$. The orthogonal basis of the feature space is defined as the eigenvectors of the total class scattering matrix,

$$\mathbf{S}_t = \frac{1}{N} \sum_{n=1}^{N} (\mathbf{x}_n - \bar{\mathbf{x}})(\mathbf{x}_n - \bar{\mathbf{x}})^T \qquad (1)$$

Where $\bar{\mathbf{x}} = \frac{1}{N} \sum_{n=1}^{N} \mathbf{x}_n$ is the mean vector. Here we are dealing with solving the eigenproblem,

$$\mathbf{S}_t \boldsymbol{\varphi} = \boldsymbol{\varphi} \boldsymbol{\Lambda} \qquad (2)$$

In which $\boldsymbol{\varphi}$ is a matrix that its columns are the eigenvectors of $\mathbf{S}_t$ respectively, and $\boldsymbol{\Lambda}$ is a diagonal matrix consisting of all the eigenvalues $\{\lambda_n \mid n = 1, 2,...,N\}$ of $\mathbf{S}_t$ along its principal diagonal, while its other elements are zero.

Practically, the algorithm proceeds by first computing the mean of $\mathbf{x}_n$ vectors and then subtracting them from this mean value. The total class scattering matrix or the covariance matrix is then calculated and its eigenvectors and eigenvalues are founded. The eigenvectors that correspond to the $k$ of largest eigenvalues are retained, and the input vectors $\mathbf{x}_n$ are projected onto the eigenvectors to give the components of the transformed vectors $\mathbf{z}_n$ in the $k$-dimensional space. Such projection results in a vector containing $k$ coefficients $a_1,...,a_k$. Hereafter, the vectors are represented by a linear combination of eigenvectors having the weights $a_1,...,a_k$. The dimensionality reduction error of the PCA can be evaluated as [9],

$$E_k = \frac{1}{2} \sum_{i=k+1}^{d} \lambda_i \qquad (3)$$

The above expression indicates that the introduced error $E_k$ can be eliminated by choosing an appropriate dimensionality reduction.

## 3.2. Unsupervised Neural Network

Recently, several efforts have been done to use neural networks for the feature generation and feature selection. A possible solution is via the so-called auto-associative networks. The employed network has $d$ input and $d$ output nodes and a single hidden layer with $k$ nodes and linear activations. During the training phase, the desired outputs are the same as the inputs. That is,

$$E_i = \sum_{j=1}^{d} (\hat{x}_j(i) - x_j(i))^2 \qquad (4)$$

Such a network has a unique minimum and the hidden layer maps the input $d$-dimensional space onto the output $k$-dimensional subspace. This procedure is equivalent to the linear PCA [10, 11]. A much better compression can be obtained by exploiting the Cybenko theorem [12]. It states that a three-layer neural network with $n$ input neurons, nonlinear transfer functions in the second layer, and linear transfer functions in the third layer of $r$ neurons can approximate any continuous function from $R^d$ to $R^k$, with an arbitrary accuracy. This is true provided that the number of neurons in the second layer is sufficiently large. Then to construct a structure, in order to compress the information, a three-layer MLP (Multi Layer Perceptron) is needed. Since the target of this network is not specific, another MLP is required to cascade this network to expand the compressed data and to make an approximation of the input. In this case, a five-layer MLP is formed in which the desired outputs are the same as the inputs. Figure 1 illustrates the architecture of a feed-forward auto-associative neural network with five layers. The outputs of third layer are the compressed data and are called the nonlinear principal components which may be viewed as a nonlinear generalization of the principal components.

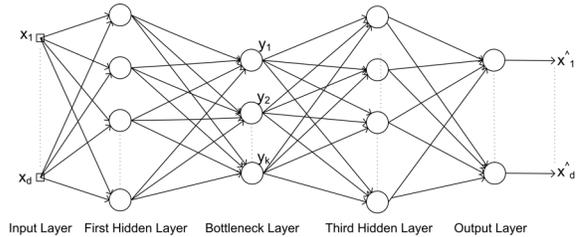

**Figure 1. A feed-forward auto-associative neural network**

The first step of using a neural network is the training phase in which the back-propagation algorithm is used. In implementation of such algorithm, two distinct passes of computation may be distinguished. The former is referred to as the forward pass, and the latter as the backward pass. In the forward pass, the synaptic weights remain unaltered throughout the network, and the function signals of the network are computed on a neuron-by-neuron basis. The Backward pass, on the other hand, starts at the output layer by passing the error signals leftward through the network, in a layer-by-layer manner. Weights and biases are updated iteratively until the MSE is minimized and the output approximates the input as closely as possible.

## 4. The Proposed Method

The BFD of computer files can be used for the content-based file type detection. However, the multiplicity of features causes to fall dawn the speed and accuracy of the recognition. Hence, a feature extraction method is needed to make a new dataset with the smaller dimensions. The PCA is a fast procedure, but it is poor when compared to the nonlinear mapping methods. On the other hand, using an auto-associative neural network with nonlinear transfer functions has a good performance. However, the computational load and the risk of being trapped in the local minima reduce its efficiency. To solve such problems, we use a hierarchical feature extraction method that serves PCA and an auto-associative neural network in order to simultaneously improve the speed and accuracy.

In the training phase, after normalizing the dataset that is obtained from the BFD, the PCA projection matrix is calculated. The covariance matrix of the training dataset is firstly calculated, as explained in section 3.1. Then the corresponding eigenvectors of the $N_1$ largest eigenvalues are selected to construct the projection matrix. Value of $N_1$ is specified according to the allowable error that is determined by (3).

The output features from the PCA are fed to an auto-associative neural network that its architecture was previously described. The number of neurons at the first and fifth layers is equal to $N_1$. The number of neurons at the third layer that is referred to as the bottleneck layer is equal to $N_2$ (where $N_2 < N_1$). Such $N_2$ features will be used for the file type detection at the classification step. This neural network is trained by using the back-propagation algorithm so that the desired outputs are the same as the inputs.

In the detection phase, the trained hierarchical feature extraction system consists of the PCA and an auto-associative neural network that provides the extracted features from the normalized BFD of the examined files. While after an effective feature extraction any classifier can be used for the file type detection, we used another three layer MLP for detecting the file type. Figure 2 shows the flowchart of the detection phase.

Although the feature extraction and classification processes can be performed by using one MLP, the above-mentioned strategy enhances the learning rate of networks and makes the turnover surveying easy. Number of neurons at the input layers is $N_2$ while it equals with the number of classes at the output layers. In the training phase of neural networks, the problem of trapping in the local minima can be avoided by introducing small disturbances on the weights after each epoch and evaluating and checking the results, whereby the step size or the learning rate can be controlled.

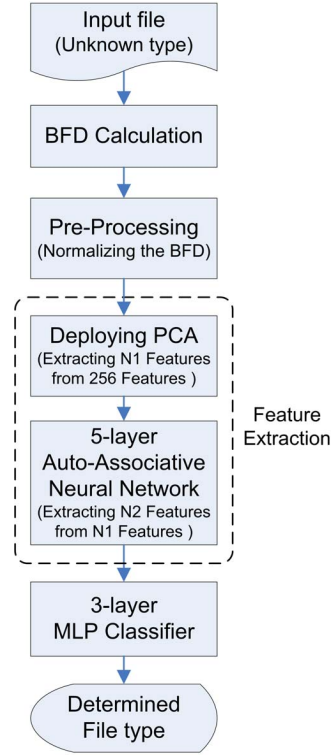

**Figure 2. Detection Phase**

## 5. Experimental Results

Our experiments was concentrated on six file types: ".doc," ".pdf," ".exe," ".jpg," ".htm" and ".gif" that are the most common file types on the Internet. The test files used in the experiments were collected from the Internet using a general search on the *Google* website. For example, the ".doc" files were collected from the *Google* search engine by using the search term ".doc". The test files can be considered as randomly chosen samples since they are captured in the wild. For the case of ".doc," ".pdf," and ".htm" files where the BFD of sample files may be language-dependent, we tried to collect some files from different languages to give more randomization and to make our results language-independent. The sizes of our sample files were completely random. While this can make our results worse, we did this to show that our results are obtained from the worst situations. We have done every thing to bother the situations so the obtained results are the results of the worst situation and the actual results will be better than the results presented here.

We collected 120 files of each type where we used 90 files out of them for training and the remained 30 files for testing the results. The initial features are the normalized BFD. The PCA reduces the number of features from 256 to $N_1$. Figure 3 shows the introduced error of the PCA for a variety number of selected features. It indicates that for the case of $N_1 = 60$, the introduced error is negligible. We repeated the procedure with different values of $N_1$ and deduced that $N_1 = 60$ is an optimum choice that can establish a good trade-off between the accuracy and computational costs.

A 5-layer auto-associative neural network with $N_2$ neurons at its bottleneck layer is trained to extract the most effective features from 60 selected features that were selected by the PCA. We reached to $N_2 = 15$ with a try-and-error experiment so we actually used 15 features to produce the fileprint of each file type. To illustrate the effectiveness of the feature extraction, Figures 4 and 5 illustrates two and three-dimensional scatter plots where the auto-associated neural network has two and three neurons at its bottleneck layer, respectively. In other words, Figures 4 and 5 are obtained by using only two and three selected features, respectively.

We used another supervised 3-layer MLP for the type detection by using the extracted features from the BFD. Such neural network has 25 neurons at its hidden layer and is trained by the back-propagation algorithm. Since only 15 features out of 256 original features are used as the fileprint of each file type, the detection speed will be very fast. The 30 remained files of each class are used for the system test. They are fed to the hierarchical feature extraction system and are then classified by the 3-layer MLP neural network. Table 1 shows the resulted confusion matrix. It indicates that all files of types ".doc," ".htm," ".jpg," and ".pdf" are correctly detected. Only two files of ".exe" class and one file of ".gif" class are incorrectly detected as ".pdf". Total correct classification rate is 98.33% that is an excellent result when compared with the results of [4] as 82%, and the results of a simple BFA [3] as 27.5%. The average accuracy rate of 82% in [4] is obtained by considering just five file types our results is obtained by considering six file types.

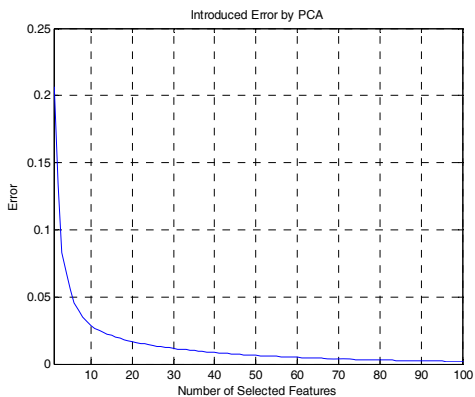

**Figure 3. Introduced error of the PCA**

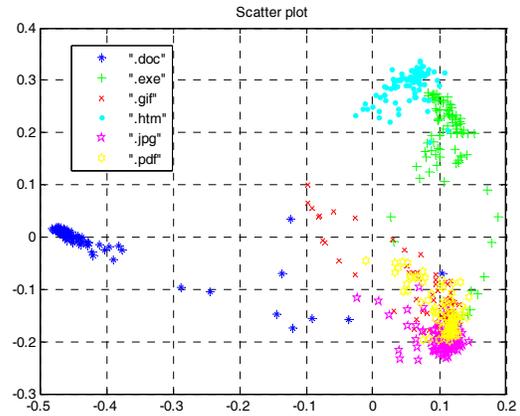

**Figure 4. Two Dimensional Scatter Plot**

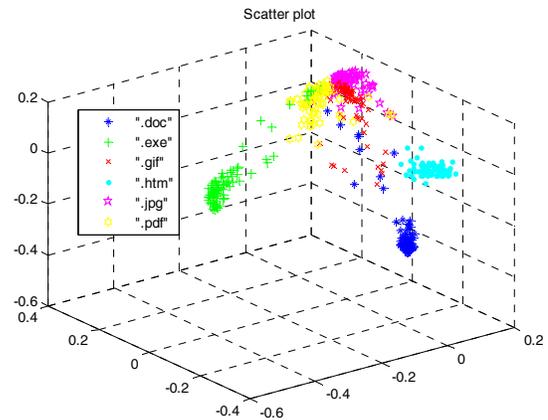

**Figure 5. Three Dimensional Scatter Plot**

**Table 1. The resulted confusion matrix for 30 examined files of each type**

|     | doc | exe | gif | htm | jpg | Pdf |
|-----|-----|-----|-----|-----|-----|-----|
| doc | 30  | 0   | 0   | 0   | 0   | 0   |
| exe | 0   | 28  | 0   | 0   | 0   | 0   |
| gif | 0   | 0   | 29  | 0   | 0   | 0   |
| htm | 0   | 0   | 0   | 30  | 0   | 0   |
| jpg | 0   | 0   | 0   | 0   | 30  | 0   |
| pdf | 0   | 2   | 1   | 0   | 0   | 30  |

It is noteworthy that we did not use file headers and file extensions and our method is completely dependent on the whole contents of files so the proposed method can work for the case of data fragments. Our results may be compared with the results of file type identification of data fragments in [6] where the basic correct detection rate is 87.3%, and it is improved to 92.1% by using their conventional RoC method [6].

While our proposed method is based on the BFD of the whole contents of files, it is not under influence of the file headers or the positions of data fragments in the files. It cannot be spoofed by changing the file header. It works even if the file header has been corrupted. It is invulnerable to the file scrambling, i.e. it cannot be spoofed by mixing and changing the positions of data fragments. It does not care the file extensions so the extension spoofing cannot take place.

Although we concentrated on six file types, the proposed method can be also used for the further types. In this case, if the confusion among the examined files were great, an interclass clustering in the confused files can be performed. In other words, such files can be assumed as two or more kinds and if the classifier detects each of them, the corresponded file type is declared. This is equivalent to the multi-centroid approach.

The correct classification rate of 98.33% is obtained from the worst situation. We considered the whole contents of files and did not do the truncation that could make our results header-dependent. We did not categorize the downloaded sample files according to their sizes. The file sizes were varying between 43 Bytes and 24.3MB and we did not care the problem of sample bias that some literatures (such as [4] and [7]) tried to prevent it by file size categorizing of the examined files. The actual results may be better than the results presented here. Table 2 shows the variation range of file sizes among our 120 downloaded sample files of each type.

**Table 2. Size variation among the 120 sample files of each type**

| Type of sample files | Maximum Size (Bytes) | Minimum Size (Bytes) |
|---|---|---|
| doc | 6906880 | 15360 |
| exe | 24265736 | 882 |
| gif | 298235 | 43 |
| htm | 705230 | 1866 |
| jpg | 946098 | 481 |
| pdf | 10397799 | 12280 |

## 6. Conclusions

A new content-based file type detection method is introduced in this paper that uses PCA and unsupervised neural networks for the automatic feature extraction. A fileprint of each file type is obtained, according to the extracted features. When an unknown file is examined, its fileprint is produced and compared with the prior produced fileprints. The proposed method has a very good accuracy and is fast enough to be deployed in real-time applications. It is completely header-independent and uses the whole contents of files. It does not depend on the positions of data fragments of the files and can detect the file type even if the file header is changed or corrupted. A total correct classification rate of 98.33% is obtained when considering the whole contents of files and without being file size specific. The proposed method can be optimized by taking several approaches. The accuracy can be improved by taking the multi-centroid models when dealing with the huge number of file types. Truncation can also improve the accuracy and the speed but it can make the method header-dependent. A file size categorization can also improve the accuracy. We will consider such viewpoints in our future papers.